\documentclass{article}

\usepackage[preprint]{neurips_2022}

\usepackage[utf8]{inputenc} 
\usepackage[T1]{fontenc}    
\usepackage{hyperref}       
\usepackage{url}            
\usepackage{booktabs}       
\usepackage{amsfonts}       
\usepackage{nicefrac}       
\usepackage{microtype}      
\usepackage{xcolor}         
\usepackage{amsmath}
\usepackage{amssymb}
\usepackage{mathtools}
\usepackage{amsthm}
\usepackage{multirow}
\usepackage{subcaption}
\usepackage{comment}
\DeclareMathOperator*{\argmax}{arg\,max}
\DeclareMathOperator*{\argmin}{arg\,min}

\newcommand{\norm}[1]{\left\lVert #1 \right\rVert}

\makeatletter
\newcommand{\subalign}[1]{%
  \vcenter{%
    \Let@ \restore@math@cr \default@tag
    \baselineskip\fontdimen10 \scriptfont\tw@
    \advance\baselineskip\fontdimen12 \scriptfont\tw@
    \lineskip\thr@@\fontdimen8 \scriptfont\thr@@
    \lineskiplimit\lineskip
    \ialign{\hfil$\m@th\scriptstyle##$&$\m@th\scriptstyle{}##$\hfil\crcr
      #1\crcr
    }%
  }%
}
\makeatother

\title{Variational Diffusion Auto-encoder: \\ Latent Space Extraction from Pre-trained Diffusion Models}

\author{
  Georgios Batzolis \thanks{Equal contibution.} \\
  DAMTP\\
  University of Cambridge\\
  Cambridge CB3 0WA \\
  \texttt{gb511@cam.ac.uk} \\
   \And
   Jan Stanczuk $ ^\ast$ \\
   DAMTP \\
   University of Cambridge \\
   Cambridge CB3 0WA \\
   \texttt{js2164@cam.ac.uk} \\
   \AND
   Carola-Bibiane Sch\"{o}nlieb \\
   DAMTP \\
   University of Cambridge \\
   Cambridge CB3 0WA \\
   \texttt{cbs31@cam.ac.uk}
}

\begin{document}

\maketitle
\begin{abstract}
As a widely recognized approach to deep generative modeling, Variational Auto-Encoders (VAEs) still face challenges with the quality of generated images, often presenting noticeable blurriness. This issue stems from the unrealistic assumption that approximates the conditional data distribution, $p(\textbf{x} | \textbf{z})$, as an isotropic Gaussian. In this paper, we propose a novel solution to address these issues. We illustrate how one can extract a latent space from a pre-existing diffusion model by optimizing an encoder to maximize the marginal data log-likelihood. Furthermore, we demonstrate that a decoder can be analytically derived post encoder-training, employing the Bayes rule for scores. This leads to a VAE-esque deep latent variable model, which discards the need for Gaussian assumptions on $p(\textbf{x} | \textbf{z})$ or the training of a separate decoder network. Our method, which capitalizes on the strengths of pre-trained diffusion models and equips them with latent spaces, results in a significant enhancement to the performance of VAEs.
\end{abstract}

\section{Introduction}
Variational Autoencoders (VAEs) \cite{vae} have proven to be a powerful tool for unsupervised learning, allowing for the efficient modeling and generation of complex data distributions. However, VAEs have important limitations, including difficulty in capturing the underlying structure of high-dimensional data and generating blurry images \cite{zaho2017understanding_vaes}. These problems emerge due to an unrealistic modeling assumption that the conditional data distribution $p(\textbf{x}| \textbf{z})$ can be approximated as a Gaussian distribution. Moreover, instead of sampling from $p(\textbf{x}| \textbf{z})$ the model simply outputs the mean of the distribution, which results in an undesirable smoothing effect. In this work we propose to relax this limiting assumption by modeling $p(\textbf{x}| \textbf{z})$ in a flexible way by leveraging the capabilities of diffusion models. We show that an encoder network modeling $p( \textbf{z} | \textbf{x})$ can be easily combined with an unconditional diffusion model trained on $p(\textbf{x})$ to yield a model for $p(\textbf{x} | \textbf{z})$.

Diffusion models \cite{diffusion_models, ddpm} have recently emerged as a promising technique for generative modeling, which uses the time reversal of a diffusion process, to estimate the data distribution $p(\textbf{x})$. Diffusion models proved to be incredibly successful in capturing complex high-dimensional distributions achieving state-of-the-art performance in many tasks such as image synthesis \cite{dhariwal2021diffusion_beats_gans} and audio generation \cite{kong2020diffWave}. Recent works show that diffusion models can capture effectively complex conditional probability distributions and apply them to solve problems such as in-painting, super resolution or image-to-image translation \cite{batzolis2022non_uniform, saharia2021sr3}.

Motivated by the success of diffusion models in learning conditional distributions, recent works \cite{preechakul2022diffusion_decoder, yang2023ldiffusion_decoder_compression} have explored applications of \textbf{conditional} diffusion models as a decoders in a VAE framework by training them to learn $p(\textbf{x} | \textbf{z})$. We improve upon this line of research by  showing that the diffusion decoder is obsolete. Instead one can combine the encoder with an \textbf{unconditional} diffusion model via the Bayes' rule for score functions to obtain a model for $p(\textbf{x}| \textbf{z})$.
This approach has several important advantages

\begin{enumerate}
    \item It avoids making unrealistic Gaussian assumption on $p(\textbf{x} | \textbf{z})$. Therefore significantly improves the performance compared to original VAE avoiding blurry samples.
    \item Since the diffusion model used in our approach is \textbf{unconditional}, i.e. it does not depend on the latent factor, our method can leverage existing powerful pre-trained  diffusion models and combine them with any encoder network. Moreover the diffusion component can be always easily replaced for a better model without the need to retrain the encoder. This is in contrast to prior approaches which used \textbf{conditional} diffusion models, which have to be trained specifically to a given encoder.
    \item By using the Bayes' rule for score functions we can separate training the prior from training the encoder and improve the training dynamics. This allows our method to achieve performance superior to approaches based on conditional diffusion models.
\end{enumerate}
Moreover we derive a novel lower-bound on the data likelihood $p(\textbf{x})$, which can be used to optimise the encoder in this framework.

We showcase the performance of our method on CIFAR10 \cite{cifar} and CelebA \cite{liu2015celeba} datasets.

\section{Background}
\subsection{Variational Autoencoders}
Variational Autoencoders (VAEs) \cite{vae, rezende2014vae2} is a deep latent variable model which approximates the data generating distribution $p(\textbf{x})$. In VAEs it is assumed that the data generating process can be represented by first generating a latent variable $z$ according to $p( \textbf{z})$ and then generating the corresponding data point $x$ according to the conditional data distribution $p(\textbf{x}| \textbf{z})$. The model is parameterized by two neural networks: the encoder $e_\phi$ and the decoder $d_\theta$. VAEs make the following parametric assumptions:
\begin{itemize}
    \item The latent prior distribution $p( \textbf{z})$ is assumed to be a standard Gaussian $\mathcal{N}(0, I)$.
    \item The data posterior distribution $p_\theta(\textbf{x} | \textbf{z})$  is parameterised as a Gaussian  $\mathcal{N}(\mu^x_\theta(\textbf{x}), \Sigma^x_\theta(\textbf{x}))$ where the mean $\mu^x_\theta(\textbf{x})$ and the diagonal covariance matrix $\Sigma^x_\theta(\textbf{x})$ are outputs of the decoder network $d_\theta: \textbf{z} \mapsto (\mu^x_\theta(\textbf{x}), \Sigma^x_\theta(\textbf{x}))$. However in many implementations an additional simplification is made: $\Sigma^x_\theta(\textbf{x})$ is assumed to be isotropic $\sigma^x_\theta I$ or even the identity matrix $I$.
\end{itemize}
This induces latent posterior $p_\theta( \textbf{z} | \textbf{x})$ and marginal likelihood distributions $p_\theta(\textbf{x})$ which are both intractable. Therefore VAEs introduce a variational approximation
\begin{itemize}
    \item The latent posterior distribution $p_\theta( \textbf{z} | \textbf{x})$ is approximated with a variational distribution $q_\phi( \textbf{z} | \textbf{x})$ which is parameterised as a Gaussian  $\mathcal{N}(\mu^z_\phi(\textbf{x}), \Sigma^z_\phi(\textbf{x}))$ where the mean $\mu^z_\phi(\textbf{x})$ and the diagonal covariance matrix $\Sigma^z_\phi(\textbf{x})$ are outputs of the encoder network $e_\phi: \textbf{x} \mapsto (\mu^z_\phi(\textbf{x}), \Sigma^z_\phi(\textbf{x}))$.
\end{itemize}

As mentioned before the exact log-likelihood $\ln p_{\theta}(\textbf{x})$ is intractable, however using the variational distribution $q_\phi( \textbf{z} | \textbf{x})$ one obtains a tractable lower bound known as the evidence lower bound (ELBO) or variational lower bound:
\begin{gather*}
    \ln p_\theta(\textbf{x}) - D_{KL}( q_\phi( \textbf{z} | \textbf{x}) \parallel p_\theta( \textbf{z} | \textbf{x})) =  \mathbb{E}_{z \sim q_\phi( \textbf{z} | \textbf{x})}[\ln p_\theta(\textbf{x} | \textbf{z})] - D_{KL}( q_\phi( \textbf{z} | \textbf{x}) \parallel p( \textbf{z}))
\end{gather*}
Finally the model can be trained by maximizing the ELBO over all data points or by minimizing:
\begin{gather}
\label{eq:elbo}
    \mathcal{L}_{\text{ELBO}}(\theta, \phi) := -\mathbb{E}_{x \sim p(\textbf{x})} \big[ \mathbb{E}_{z \sim q_\phi( \textbf{z} | \textbf{x})}[\ln p_\theta(\textbf{x} | \textbf{z})] - D_{KL}( q_\phi( \textbf{z} | \textbf{x}) \parallel p( \textbf{z})) \big]
\end{gather}
via a SGD based optimization method.
This has the effect of both maximizing the data log-likelihood $\ln p_\theta(\textbf{x})$ and minimizing $D_{KL}( q_\phi( \textbf{z} | \textbf{x}) \parallel p_\theta( \textbf{z} | \textbf{x}))$, and therefore pushing the variational distribution $q_\phi( \textbf{z} | \textbf{x})$ towards the posterior $ p_\theta( \textbf{z} | \textbf{x})$.

It is a well known problem that the KL penalty term in \ref{eq:elbo} leads to significant over-regularization and poor convergence. Therefore many implementations use the following modified $\beta$-ELBO objective instead \cite{higgins2016beta_vae}:
\begin{gather}
\label{eq:elbo}
    \mathcal{L}_{\text{ELBO}}(\theta, \phi, \beta) := -\mathbb{E}_{x \sim p(\textbf{x})} \big[ \mathbb{E}_{z \sim q_\phi( \textbf{z} | \textbf{x})}[\ln p_\theta(\textbf{x} | \textbf{z})] - \beta D_{KL}( q_\phi( \textbf{z} | \textbf{x}) \parallel p( \textbf{z})) \big]
\end{gather}
where $\beta \in (0,1)$. The resulting model is often referred to as $\beta$-VAE \footnote{It is worth noting that choosing $\beta \not = 1$ is equivalent to choosing a fixed isotropic covariance $\Sigma^x_\theta(\textbf{x}) = \sigma I$ for appropriate value of $\sigma$, instead of the identity matrix. We refer to \cite{rybkin2021sigma_vae} for details}.

\subsection{Score-based diffusion models}
\label{sec:background_score}

\textbf{Setup:} In \cite{song2020score} score-based  \cite{score_matching} and diffusion-based \cite{diffusion_models, ddpm} generative models have been unified into a single continuous-time score-based framework where the diffusion is driven by a stochastic differential equation.  This framework relies on Anderson's Theorem \cite{anderson1982reverse_time_sde}, which states that under certain Lipschitz conditions on the drift coefficient $f : \mathbb{R}^{n_x} \times \mathbb{R} \xrightarrow{} \mathbb{R}^{n_x}$ and on the diffusion coefficient $G : \mathbb{R}^{n_x} \times \mathbb{R}\xrightarrow{} \mathbb{R}^{n_x} \times \mathbb{R}^{n_x}$ and an integrability condition on the target distribution $p(\textbf{x}_0)$ a forward diffusion process governed by the following SDE:
\begin{gather}
\label{eq:forward_sde}
 d\textbf{x}_t = f(\textbf{x}_t,t)dt+G(\textbf{x}_t,t)d\textbf{w}_t  
\end{gather} 
has a reverse diffusion process governed by the following SDE:
\begin{gather}\label{eq:reverse_sde}
d\textbf{x}_t=[f(\textbf{x}_t,t)-G(\textbf{x}_t,t)G(\textbf{x}_t,t)^T\nabla_{\textbf{x}_t}{\ln{p_{\textbf{X}_t}(\textbf{x}_t)}}]dt + G(\textbf{x}_t,t)d\Bar{\textbf{w}_t},
\end{gather}

\noindent where $\Bar{\textbf{w}_t}$ is a standard Wiener process in reverse time. 

The forward diffusion process transforms the \textit{target distribution} $p(\textbf{x}_0)$ to a \textit{diffused distribution} $p(\textbf{x}_T)$ after diffusion time $T$. By appropriately selecting the drift and the diffusion coefficients of the forward SDE, we can make sure that after sufficiently long time $T$, the diffused distribution $p(\textbf{x}_T)$ approximates a simple distribution, such as $\mathcal{N}(\textbf{0},\textbf{I})$. We refer to this simple distribution as the \textit{prior distribution}, denoted by $\pi$. The reverse diffusion process transforms the diffused distribution $p(\textbf{x}_T)$ to the data distribution $p(\textbf{x}_0)$ and the prior distribution $\pi$ to a distribution $p^{SDE}$. The distribution $p^{SDE}$ is close to $p(\textbf{x}_0)$ if the diffused distribution $p(\textbf{x}_T)$ is close to the prior distribution $\pi$. We get samples from $p^{SDE}$ by sampling from $\pi$ and simulating the reverse SDE from time $T$ to time $0$.

\textbf{Sampling:} To get samples by simulating the reverse SDE, we need access to the time-dependent \textit{score function} $\nabla_{\textbf{x}_t}{\ln{p(\textbf{x}_t)}}$. In practice, we approximate the time-dependent score function with a neural network $s_{\theta}(\textbf{x}_t,t) \approx \nabla_{\textbf{x}_t}{\ln{p(\textbf{x}_t)}}$ and simulate the reverse SDE presented in equation \ref{eq:approximated_reverse_sde} to map the prior distribution $\pi$ to $p^{SDE}_{\theta}$.

\begin{gather}\label{eq:approximated_reverse_sde}
d\textbf{x}_t=[f(\textbf{x}_t,t)-G(\textbf{x}_t,t)G(\textbf{x}_t,t)^Ts_{\theta}(\textbf{x}_t,t)]dt + G(\textbf{x}_t,t)d\Bar{\textbf{w}_t},
\end{gather}If the prior distribution is close to the diffused distribution and the approximated score function is close to the ground truth score function, the modeled distribution  $p^{SDE}_{\theta}$ is provably close to the target distribution $p(\textbf{x}_0)$. This statement is formalised in the language of distributional distances in the work of \cite{song2021maximum}. 




\textbf{Training:} A neural network $s_\theta(\textbf{x}_t,t)$ can be trained to approximate the score function $\nabla_{\textbf{x}_t}{\ln{p(\textbf{x}_t)}}$ by minimizing the weighted score matching objective

\begin{gather}
\begin{aligned}
    &\mathcal{L}_{SM}(\theta, \lambda(\cdot)) := 
    &\frac{1}{2} \mathbb{E}_{\subalign{t \sim U(0,T)\\ \textbf{x}_t \sim p(\textbf{x}_t)}} [\lambda(t) \norm{\nabla_{\textbf{x}_t}{\ln{p(\textbf{x}_t)}} - s_\theta(\textbf{x}_t,t)}_2^2]
\end{aligned}
\end{gather}
where $\lambda: [0,T] \xrightarrow{} \mathbb{R}_+$ is a positive weighting function.

However, the above quantity cannot be optimized directly since we don't have access to the ground truth score $\nabla_{\textbf{x}_t}{\ln{p(\textbf{x}_t)}}$. Therefore in practice, a different objective has to be used \cite{score_matching, vincent2011connection, song2020score}. In \cite{song2020score}, the weighted denoising score-matching objective is used, which is defined as 

\begin{gather}\label{DSM for uniform diffusion models}
\begin{aligned}
    &\mathcal{L}_{DSM}(\theta, \lambda(\cdot)) := 
    &\frac{1}{2} \mathbb{E}_{\subalign{t \sim U(0,T)\\ \textbf{x}_0 \sim p(\textbf{x}_0) \\ \textbf{x}_t \sim p(\textbf{x}_t | \textbf{x}_0)}} [\lambda(t) \norm{\nabla_{\textbf{x}_t}{\ln{p(\textbf{x}_t | \textbf{x}_0)}} - s_\theta(\textbf{x}_t,t)}_2^2]
\end{aligned}
\end{gather}

The difference between DSM and SM is the replacement of the ground truth score which we do not know by the score of the perturbation kernel which we know analytically for many choices of forward SDEs. The choice of the weighted DSM objective is justified because the weighted DSM objective is equal to the SM objective up to a constant that does not depend on the parameters of the model $\theta$. The reader can refer to \cite{vincent2011connection} for the proof. 

\textbf{Weighting function:} The choice of the weighting function is also important, because it determines the quality of score-matching in different diffusion scales. In the case of \textit{uniform diffusion} i.e.  when $G(\textbf{x}_t,t) = g(t) I$ for $g: \mathbb{R} \xrightarrow{} \mathbb{R}$ a principled choice for the weighting function is $\lambda(t) = g(t)^2$. This weighting function is called the likelihood weighting function \cite{song2021maximum}, because it ensures that we minimize an upper bound on the Kullback–Leibler divergence from the target distribution to the model distribution by minimizing the weighted DSM objective with this weighting. The previous statement is implied by the combination of inequality \ref{Likelihood Weighting for Uniform Diffusion Models} which is proven in \cite{song2021maximum} and the relationship between the DSM and SM objectives.

\begin{equation}\label{Likelihood Weighting for Uniform Diffusion Models}
\begin{aligned}
    D_{KL}(p(\textbf{x}_0)\parallel p^{SDE}_{\theta}(\textbf{x}_0)) &\leq    \mathcal{L}_{SM}(\theta, g(\cdot)^2) & + D_{KL}(p(\textbf{x}_T)\parallel \pi)
\end{aligned}
\end{equation}

Other weighting functions have also yielded very good results \cite{kingmaVDM} with particular choices of forward SDEs. However, we do not have theoretical guarantees that alternative weightings would yield good results with arbitrary choices of forward SDEs.

\textbf{Conditional diffusion models:} The continuous score-matching framework can be extended to conditional generation, as shown in  \cite{song2020score}. Suppose we are interested in $p(\textbf{x} | \textbf{z})$, where $\textbf{x}$ is a \textit{target data} and $\textbf{z}$ is a \textit{condition}. Again, we use the forward diffusion process (Equation \ref{eq:forward_sde}) to obtain a family of diffused distributions $p(\textbf{x}_t | \textbf{z})$ and apply Anderson's Theorem to derive the \textit{conditional reverse-time SDE}
\begin{equation}
\label{eq:conditional_reverse_sde}
    dx = [f(\textbf{x}_t,t) -G(\textbf{x}_t,t)G(\textbf{x}_t,t)^T\nabla_{\textbf{x}_t}{\ln{p(\textbf{x}_t | \textbf{z})}}]dt + G(\textbf{x}_t,t)d\Bar{\textbf{w}_t}.
\end{equation}
Now we need to learn the conditional score $\nabla_{\textbf{x}_t} \ln{p(\textbf{x}_t | \textbf{z})}$ in order to be able to sample from $p(\textbf{x} | \textbf{z})$ using reverse-time diffusion.

The conditional denoising estimator (CDE) is a way of estimating $\nabla_{\textbf{x}_t} \ln{p(\textbf{x}_t | \textbf{z})}$ using the denoising score matching approach \cite{vincent2011connection, song2020score}. In order to approximate $\nabla_{\textbf{x}_t} \ln{p(\textbf{x}_t | \textbf{z})}$, the conditional denoising estimator minimizes
\begin{gather}
\begin{aligned}
        \label{CDE}
        &\frac{1}{2} \mathbb{E}_{\subalign{&t \sim U(0,T)\\ &x_0, z \sim p(\textbf{x}_0, z) \\ &x_t \sim p(\textbf{x}_t | \textbf{x}_0)}} 
        [\lambda(t) \norm{\nabla_{\textbf{x}_t}{\ln{p(\textbf{x}_t | \textbf{x}_0)}} -  s_\theta(\textbf{x}_t, \textbf{z},t)}_2^2].
\end{aligned}
\end{gather}
In \cite{batzolis2022non_uniform} it has been shown that this is equivalent to minimizing 
\begin{gather*}
        \frac{1}{2} \mathbb{E}_{\subalign{&t \sim U(0,T)\\ &x_t, z \sim p(\textbf{x}_t, z)}} 
        [\lambda(t) \norm{\nabla_{\textbf{x}_t} \ln{p(\textbf{x}_t | \textbf{z})} - s_\theta(\textbf{x}_t, \textbf{z},t)}_2^2]
\end{gather*}
and that under mild assumptions $ s_\theta(\textbf{x}_t, \textbf{z},t)$ is a consistent estimator of the conditional score $\nabla_{\textbf{x}_t} \ln{p(\textbf{x}_t | \textbf{z})}$.

\section{Method}

\begin{figure}
    \centering
    \includegraphics[width=0.95\textwidth]{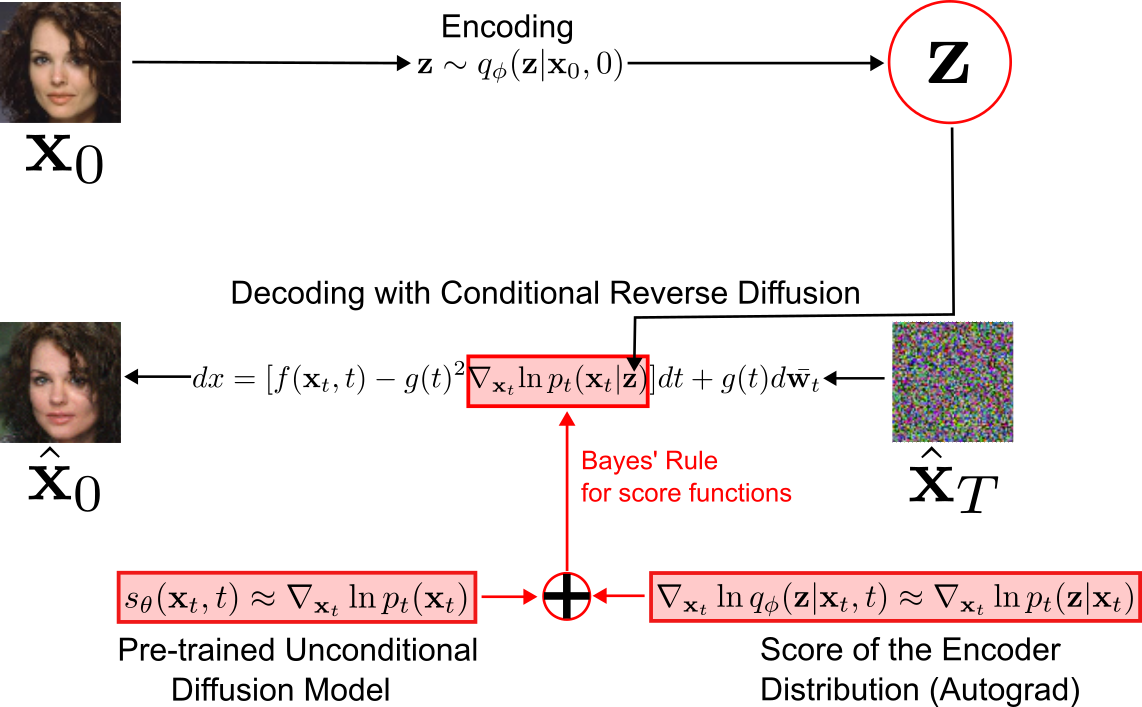}
    \caption{Graphical overview of our method. The time-dependent encoder network $e_\phi$ induces the encoder distribution $q_\phi( \textbf{z} | \textbf{x}_t, t) \approx p_t( \textbf{z} | \textbf{x}_t)$. The data $\textbf{x}_0$ is encoded with the encoder  into a latent vector $\textbf{z}$ by sampling $q_\phi( \textbf{z} | \textbf{x}_0, 0)$. Then the reconstruction $\hat{\textbf{x}}_0$ is obtained by running the conditional reverse diffusion process using the approximate conditional data score $s_{\theta, \phi}(\textbf{x}_t, \textbf{z},t) \approx \nabla_{\textbf{x}_t}  \ln{p(\textbf{x}_t | \textbf{z})}$. The model $s_{\theta, \phi}(\textbf{x}_t, \textbf{z},t)$ is obtained by adding the score of unconditional diffusion  model $s_\theta(\textbf{x}_t,t) \approx \nabla_{\textbf{x}_t}  \ln{p(\textbf{x}_t)} $ and the score of the encoder distribution $ \nabla_{\textbf{x}_t}  \ln q_\phi( \textbf{z} | \textbf{x}_t, t) \approx
 \nabla_{\textbf{x}_t}  \ln{p(\textbf{z} | \textbf{x}_t )} $. The latter can be computed via automatic differentiation with respect to the input $\textbf{x}_t$.}
\end{figure}

\subsection{Problems with conventional VAEs}
As discussed before, VAEs model the conditional data distribution $p_\theta (\textbf{x}| \textbf{z})$ as a Gaussian distribution with mean $\mu^x_\theta( \textbf{z})$ and covariance $\Sigma^x_\theta( \textbf{z})$ learned by the decoder network $d_\theta$. Moreover, in practice it is often assumed that $\Sigma^x_\theta = I$. Under these assumptions maximizing the conditional log-likelihood $$\ln p_\theta (\textbf{x} | \textbf{z}) = -\frac{\norm{z - \mu^x_\theta( \textbf{z})}_2^2}{2} + \frac{d}{2}\ln{2\pi}$$ is equivalent to minimizing the $L_2$ reconstruction error. There are several reasons why this model is inadequate when dealing with certain data modalities such as images:
\begin{itemize}
    \item Samples from $p_\theta(\textbf{x} | \textbf{z})$ would look like noisy images and therefore instead of sampling the distribution $p_\theta(\textbf{x}| \textbf{z})$, as a principled model should, VAEs simply output the mean $\mu^x_\theta( \textbf{z})$, which leads to undesirable smoothing and blurry samples \cite{zaho2017understanding_vaes}.
    \item It is equivalent to $L_2$ pixel-wise error, which aligns very poorly with human perception and semantics of the image \cite{zhang2018lpips}. See appendix  \ref{sec: l2} for a detailed discussion.
\end{itemize}

\subsection{Conditional Diffusion Models as decoders}
To mitigate above problems and avoid making the unrealistic Gaussian assumption about $p(\textbf{x}| \textbf{z})$ one can train a conditional diffusion model. Such approaches have been explored in \cite{preechakul2022diffusion_decoder, yang2023ldiffusion_decoder_compression}. The conditional diffusion model $s_\theta(\textbf{x}_t, \textbf{z},t)$ is trained jointly with an encoder network $e_\phi : x_0 \mapsto z$ to approximate the conditional data score $\nabla_{\textbf{x}_t}  \ln{p(\textbf{x}_t | \textbf{z})}$ by minimizing the objective \ref{CDE}. This significantly improves upon original auto-encoder framework by avoiding the Gaussian assumption and alleviates the problem of blurry samples. However, our experiments presented in section \ref{Experiments} indicate that this framework fails when trained as a Variational Autoencoder (VAE), that is, upon the introduction of the Kullback-Leibler (KL) penalty term, even when the regularization coefficient $\beta$ is very small. For the remainder of this paper, we will refer to this training method as DiffDecoder.

\subsection{Score VAE: Encoder with unconditional diffusion model as prior}
In this work, we propose a further development to the above idea by leveraging Baye's rule for scores and using the structure of $\nabla_{\textbf{x}_t}  \ln{p(\textbf{x}_t | \textbf{z})}$. We can separate the training of the prior from training the encoder and improve the training dynamics.
By  Bayes's rule for scores we have:
\begin{equation*}
    \nabla_{\textbf{x}_t}  \ln{p(\textbf{x}_t | \textbf{z})} =  \nabla_{\textbf{x}_t}  \ln{p(\textbf{z} | \textbf{x}_t )} + \nabla_{\textbf{x}_t}  \ln{p(\textbf{x}_t)} 
\end{equation*}
This means that we can decompose the \textit{conditional data score} $\nabla_{\textbf{x}_t}  \ln{p(\textbf{x}_t | \textbf{z})} $ into the \textit{data prior score} $\nabla_{\textbf{x}_t}  \ln{p(\textbf{x}_t)} $ and the \textit{latent posterior score} $\nabla_{\textbf{x}_t}  \ln{p(\textbf{z} | \textbf{x}_t )}$. The data prior score can be approximated by an unconditional diffusion model $s_\theta(\textbf{x}_t,t)$ in the data space, allowing us to leverage powerful pretrained diffusion models. The latent posterior score $ \nabla_{\textbf{x}_t}  \ln{p(\textbf{z} | \textbf{x}_t )}$ is approximated by a time-dependent encoder network $e_\phi(\textbf{x}_t, t)$. We will discuss the details of modeling the latent posterior score in the next section. Once we have the model for data prior and latent posterior scores we can combine them to obtain the conditional data score. Then we obtain a complete latent variable model. The data $\textbf{x}$ can be encoded to latent representation $\textbf{z}$ using the encoder network and then it can by reconstructed by simulating the conditional reverse diffusion process using the conditional data score $\nabla_{\textbf{x}_t}  \ln{p(\textbf{x}_t | \textbf{z})}$.

This method has several advantages over the conditional diffusion approach, while preserving the benefits of having a powerful and flexible model for $p(\textbf{x}| \textbf{z})$.  In the conditional diffusion case the score model $s_\theta(\textbf{x}_t, \textbf{z},t)$ has to be trained \textbf{jointly} with the encoder network $e_\phi$. Moreover,  $s_\theta(\textbf{x}_t, \textbf{z},t)$  has to implicitly learn two distributions. Firstly it has to approximate $p( \textbf{z}| \textbf{x})$ to understand how $e_\phi$ encodes the information about $\textbf{x}$ into $\textbf{z}$ and secondly it has to model the prior $p(\textbf{x})$ to ensure realistic reconstructions. In our approach these two tasks are clearly separated and delegated to two separate networks. Therefore, the diffusion model does not need to re-learn the encoder distribution. Instead the prior and the encoder distributions are combined in a principled analytical way via the Bayes' rule for score functions. 

Moreover, the unconditional prior model $s_\theta(\textbf{x}_t, t)$ can be trained first, \textbf{independently} of the encoder. Then we freeze the prior model and train just the encoder network. This way we always train only one network at a time, what allows for improved training dynamics.

Additionally, the data prior network can be always replaced by a better model without the need to retrain the encoder.

\subsection{Modeling the latent posterior score $\nabla_{\textbf{x}_t}  \ln{p(\textbf{z} | \textbf{x}_t )}$}
The latent posterior score is induced by the encoder network. First similarly to VAEs we impose a Gaussian parametric model at $t=0$:
\begin{equation*}
    p_\phi(\textbf{z} | \textbf{x}_0 ) = \mathcal{N}(\textbf{z} ; \mu^z_\phi(\textbf{x}_0), \sigma^z_\phi(\textbf{x}_0)I)
\end{equation*}
where $ \mu^z_\phi(\textbf{x}_0), \sigma^z_\phi(\textbf{x}_0)$ are the outputs of the encoder network.
This together with the transition kernel $p(\textbf{x}_t| \textbf{x}_0)$ determines the distribution $p_\phi(\textbf{z} | \textbf{x}_t)$, which is given by
\begin{equation}
\label{eq:true_posterior}
    p_\phi(\textbf{z} | \textbf{x}_t) = \mathbb{E}_{x_0 \sim p(\textbf{x}_0 | \textbf{x}_t)}[p_\phi( \textbf{z} | \textbf{x}_0)]
\end{equation}
The above is computationally intractable since sampling from $ p_t(\textbf{x}_0 | \textbf{x}_t)$ would require solving the reverse SDE multiple times during each training step. Therefore we consider a variational approximation to the above distribution 
\begin{equation}\label{definition of variational approximation}
    q_{t, \phi}(\textbf{z}|\textbf{x}_t) =  \mathcal{N}(\textbf{z} ; \mu_\phi(\textbf{x}_t , t), \sigma_\phi(\textbf{x}_t, t))
\end{equation}
and learn parameters $\phi$ such that $q_{t, \phi}(\textbf{z}|\textbf{x}_t) \approx p_t(\textbf{z}|\textbf{x}_t)$.

The choice of the above variational family is justified by the following observations:
\begin{enumerate}
    \item At time $t=0$ the true distribution belongs to the family, since  $p(\textbf{z} | \textbf{x}_0)$ is Gaussian. Moreover since for small $t$ the distribution $p_t(\textbf{x}_0 | \textbf{x}_t)$ is very concentrated around $\textbf{x}_0$ it is apparent from equation \ref{eq:true_posterior} that $p_\phi(\textbf{z} | \textbf{x}_t)$ is approximately Gaussian. 
    \item At time $t=1$ the true distribution can be well approximated by a member of the variational family. This is because a noisy sample $\textbf{x}_1$ no longer contains information about $\textbf{z}$, therefore $ p_1(\textbf{z} | \textbf{x}1) \approx p(\textbf{z})$. And since we are training with KL loss $ p(\textbf{z})$ will be approximately Gaussian. 
\end{enumerate}

Finally, we can use automatic differentiation to compute $\nabla_{\textbf{x}_t} \ln q_{t, \phi}(\textbf{z} | \textbf{x}_t)  $ which is our model for the latent posterior score $\nabla_{\textbf{x}_t}  \ln{p(\textbf{z} | \textbf{x}_t )}$.

\subsection{Encoder Training Objective}\label{Encoder Training Objective}
Let $s_\theta(\textbf{x}_t, t) \approx \nabla_{\textbf{x}_t}  \ln{p(\textbf{x}_t)}  $ be a score function of a pre-trained unconditional diffusion model. Let $e_\phi: (\textbf{x}_t, t) \mapsto (\mu_\phi^z(\textbf{x}_t, t), \sigma^z_\phi(\textbf{x}_t, t)I)$ be the encoder network, which defines the variational distribution $q_{t, \phi}(\textbf{z} | \textbf{x}_t) = \mathcal{N}\big( \textbf{z} ; \mu_\phi^z(\textbf{x}_t, t), \sigma^z_\phi(\textbf{x}_t, t)I \big)$ and $\nabla_{\textbf{x}_t} \ln q_{t, \phi}(\textbf{z} | \textbf{x}_t)  $ which approximates $\nabla_{\textbf{x}_t}  \ln{p(\textbf{z} | \textbf{x}_t )}$. By the Bayes' rule for score functions the neural approximation of the conditional data score  $\nabla_{\textbf{x}_t}  \ln{p(\textbf{x}_t | \textbf{z})}$ is given by
\begin{gather*}
s_{\theta, \phi}(\textbf{x}_t, \textbf{z}, t) : = s_\theta(\textbf{x}_t, t) + \nabla_{\textbf{x}_t} \ln q_{t, \phi}(\textbf{z} | \textbf{x}_t)    
\end{gather*}
 We train the encoder by maximizing the marginal data log-likelihood $\ln p_{\theta, \phi}(\textbf{x}) $.
In Appendix \ref{sec:mle_objective} we show that minimizing the following training objective (with $\beta=1$) is equivalent to maximizing the marginal data log-likelihood $\ln p_{\theta, \phi}(\textbf{x}) $,
\begin{equation*}
\begin{aligned}
    \mathcal{L}_\beta(\phi)  := \mathbb{E}_{\textbf{x}_0 \sim p(\textbf{x}_0)} \bigg[ 
    \frac{1}{2} \mathbb{E}_{\subalign{t \sim U(0,T) \\ \textbf{x}_t \sim p_t(\textbf{x}_t | \textbf{x}_0) \\ \textbf{z} \sim q_{0,\phi}(\textbf{z} | \textbf{x}_t) }} 
    \big[
    g(t)^2  \norm{ \nabla_{\textbf{x}_t}{\ln{p_t(\textbf{x}_t | \textbf{x}_0)}} - s_{\theta, \phi}(\textbf{x}_t, \textbf{z}, t)}_2^2  
    & \big] \\
    + \beta D_{KL}\big( q_{0,\phi}(\textbf{z} | \textbf{x}_0)  \parallel p(\textbf{z})\big)  & \bigg].
\end{aligned}    
\end{equation*}

For the remainder of this paper, we will refer to this training method as ScoreVAE.

\subsection{Correction of the variational error}
Once the encoder $e_{\phi}$ is trained, our approximation of the ground truth decoding score $\nabla_{\textbf{x}_t}  \ln{p(\textbf{x}_t | \textbf{z})}$ is $s_{\theta, \phi}(\textbf{x}_t, \textbf{z}, t) : = s_\theta(\textbf{x}_t, t) + \nabla_{\textbf{x}_t} \ln q_{t, \phi}(\textbf{z} | \textbf{x}_t)$. Even in the case of perfect optimisation, our approximation will not match the ground truth decoding score because of the variational approximation described in \ref{definition of variational approximation}. We can correct the variational approximation error after training the encoder, by training an auxiliary correction model that approximates the residual. More specifically, we define our approximation of the decoding score to be the following: 
$s_{\theta, \phi}(\textbf{x}_t, \textbf{z}, t) : = s_\theta(\textbf{x}_t, t) + \nabla_{\textbf{x}_t} \ln q_{t, \phi}(\textbf{z} | \textbf{x}_t)+c_{\psi}(\textbf{x}_t, \textbf{z}, t)$ and train the corrector model $c_{\psi}$ using the same objective as in section \ref{Encoder Training Objective} after freezing the weights of the encoder and of the prior model which have already been trained. For the remainder of this paper, we will refer to this training method as ScoreVAE+.

\section{Experiments} \label{Experiments}
We trained our methods ScoreVAE and ScoreVAE+ on Cifar10 and CelebA $64\times 64$ using $\beta = 0.01$. We present a quantitative comparison of our method to a $\beta$-VAE and DiffDecoder trained with the same $\beta$ value in Tables \ref{tbl:Cifar10} and \ref{tbl:CelebA}. We present a qualitative comparison in Tables \ref{fig:cifar10 qualitative comparison}, \ref{fig:celebA qualitative comparison} and a more extensive qualitative comparison in Figures \ref{fig:extended cifar10 qualitative comparison}, \ref{fig:extended celebA qualitative comparison} in the Appendix \ref{sec:extended qualitative evaluation}. To ensure fair comparison, we designed the models for our method, DiffDecoder and $\beta$-VAE with a very similar architecture and almost the same number of parameters. Additional experimental details are in Appendix \ref{Experimental details}. 

The DiffDecoder fails to provide consistent estimates when trained with $\beta=0.01$ for both experiments. However, it produces consistent non-blurry reconstructions when trained as auto-encoder, i.e. with $\beta=0$, as also shown in previous works \cite{preechakul2022diffusion_decoder, yang2023ldiffusion_decoder_compression}. ScoreVAE outperforms both $\beta$-VAE and DiffDecoder in both experiments according to both the quantitative metrics and the qualitative results. It should be mentioned that $\beta$-VAE achieves a slightly better $L_2$ score in CelebA $64\times 64$. However, 
$\beta$-VAE is trained to specifically minimize $L_2$, which is not a reliable metric for assessing the performance of VAEs (c.f. Appendix \ref{sec: l2}). Both quantitative metrics and qualitative results demonstrate that correcting the variational error, as done in ScoreVAE+, provides only a slight improvement over the encoder-only method (ScoreVAE). This finding reinforces the suitability of the variational assumption.

\begin{table*}[h!]
\resizebox{\textwidth}{!}{
\begin{minipage}{.5\textwidth}
    \begin{center}
    \caption{Cifar10}
    \label{tbl:Cifar10}
    \begin{tabular}{cccc}
    \toprule
    & $L_2$ & LPIPS  \\
    \midrule
    \multirow{1}{*}{VAE ($\beta=0.01$)} 
    & 3.410 & 0.269     \\
    \multirow{1}{*}{ScoreVAE ($\beta=0.01$)}
    &2.634 & 0.125  \\ 
    \multirow{1}{*}{ScoreVAE+ ($\beta=0.01$)}
    &\textbf{2.591} & \textbf{0.119} \\
    \multirow{1}{*}{DiffDecoder ($\beta=0.01$)}
    &19.53 & 0.562 \\
    \midrule
    \multirow{1}{*}{DiffDecoder ($\beta=0$)}
    &2.851 & 0.127 \\
    \bottomrule
    \end{tabular}
    \end{center}
\end{minipage}
\hspace{5mm}
\begin{minipage}{.5\textwidth}
    \begin{center}
    \caption{CelebA $64\times 64$}
    \label{tbl:CelebA}
    \begin{tabular}{cccc}
    \toprule
    & $L_2$ & LPIPS  \\
    \midrule
    \multirow{1}{*}{VAE ($\beta=0.01$)} 
    & \textbf{6.97} & 0.217    \\
    \multirow{1}{*}{ScoreVAE ($\beta=0.01$)}
    &7.322 & 0.158  \\
    \multirow{1}{*}{ScoreVAE+ ($\beta=0.01$)}
    &7.248 & \textbf{0.155} \\
    \multirow{1}{*}{DiffDecoder ($\beta=0.01$)}
    &40.25 & 0.476 \\
    \midrule
    \multirow{1}{*}{DiffDecoder ($\beta=0$)}
    &8.626 & 0.166 \\
    \bottomrule
    \end{tabular}
    \end{center}
\end{minipage}}
\end{table*}

\begin{table}[h!]
    \centering
    \resizebox{0.9\textwidth}{!}{
    \begin{tabular}{cccccc}
        Original & \multicolumn{1}{c}{ScoreVAE} & \multicolumn{1}{c}{ScoreVAE+} & \multicolumn{2}{c}{DiffDecoder} & \multicolumn{1}{c}{VAE}  \\ 
        & ($\beta=0.01$) & ($\beta=0.01$) & ($\beta=0.01$) & ($\beta=0$) & ($\beta=0.01$) \\

        \includegraphics[width=.145\textwidth]{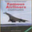} &   
        \includegraphics[width=.145\textwidth]{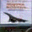} &
        \includegraphics[width=.145\textwidth]{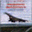} &
        \includegraphics[width=.145\textwidth]{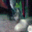} &
        \includegraphics[width=.145\textwidth]{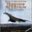} &
        \includegraphics[width=.145\textwidth]{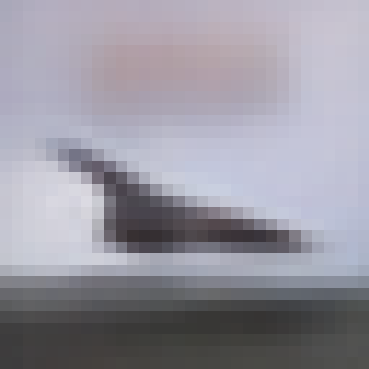} \\

        \includegraphics[width=.145\textwidth]{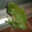} &   
        \includegraphics[width=.145\textwidth]{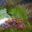} &
        \includegraphics[width=.145\textwidth]{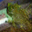} &
        \includegraphics[width=.145\textwidth]{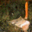} &
        \includegraphics[width=.145\textwidth]{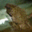} &
        \includegraphics[width=.145\textwidth]{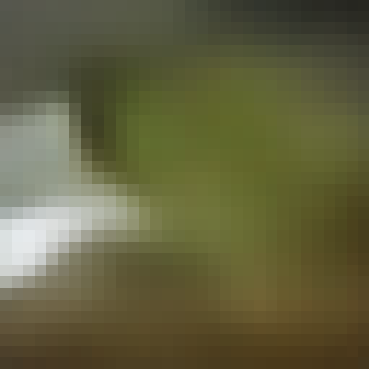} \\
 
        \includegraphics[width=.145\textwidth]{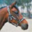} &   
        \includegraphics[width=.145\textwidth]{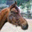} &
        \includegraphics[width=.145\textwidth]{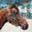} &
        \includegraphics[width=.145\textwidth]{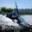} &
        \includegraphics[width=.145\textwidth]{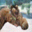} &
        \includegraphics[width=.145\textwidth]{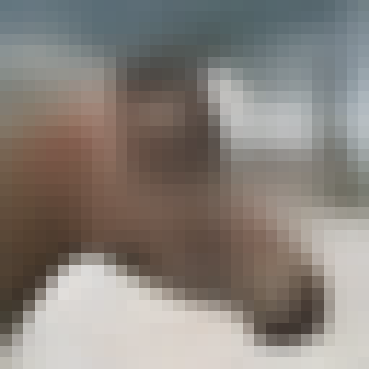} \\
    \end{tabular}}
    \caption{Cifar10}
    \label{fig:cifar10 qualitative comparison}
\end{table}
\begin{table}[h!]
    \centering
    \resizebox{0.9\textwidth}{!}{
    \begin{tabular}{cccccc}
        Original & ScoreVAE & ScoreVAE+ & \multicolumn{2}{c}{DiffDecoder} & VAE  \\ 
        & ($\beta=0.01$) & ($\beta=0.01$) & ($\beta=0.01$) & ($\beta=0$) & ($\beta=0.01$) \\

        \includegraphics[width=.15\textwidth]{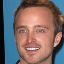} &   
        \includegraphics[width=.15\textwidth]{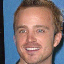} &
        \includegraphics[width=.15\textwidth]{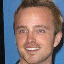} &
        \includegraphics[width=.15\textwidth]{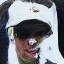} &
        \includegraphics[width=.15\textwidth]{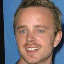} &
        \includegraphics[width=.15\textwidth] 
        {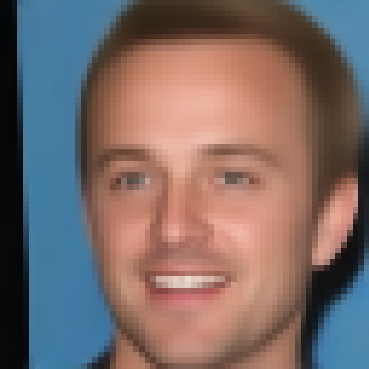} \\

        \includegraphics[width=.15\textwidth]{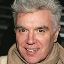} &   
        \includegraphics[width=.15\textwidth]{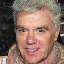} &
        \includegraphics[width=.15\textwidth]{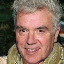} &
        \includegraphics[width=.15\textwidth]{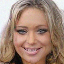} &
        \includegraphics[width=.15\textwidth]{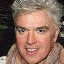} &
        \includegraphics[width=.15\textwidth]{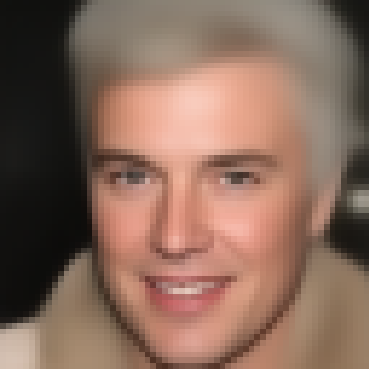} \\
 
        \includegraphics[width=.15\textwidth]{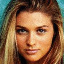} &   
        \includegraphics[width=.15\textwidth]{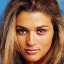} &
        \includegraphics[width=.15\textwidth]{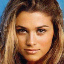} &
        \includegraphics[width=.15\textwidth]{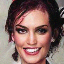} &
        \includegraphics[width=.15\textwidth]{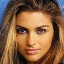} &
        \includegraphics[width=.15\textwidth]{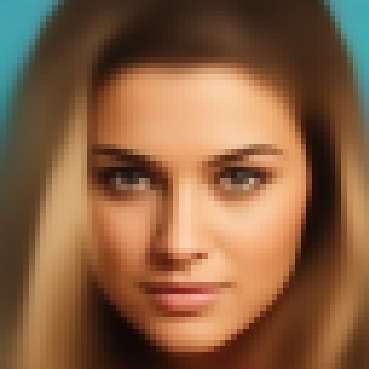} \\
    \end{tabular}}
    \caption{CelebA $64\times 64$}
    \label{fig:celebA qualitative comparison}
\end{table}

 %
 %
%

\newpage
\section{Conclusions}

In this paper, we introduce a technique that enhances the Variational Auto-Encoder (VAE) framework by employing a diffusion model, thus bypassing the unrealistic Gaussian assumption inherent in the conditional data distribution $p(\textbf{x} | \textbf{z})$. We demonstrate using the Bayes' rule for score functions that an encoder, when paired with a pre-trained unconditional diffusion model, can result in a highly effective model for $p(\textbf{x}| \textbf{z})$. Thus, we show that  provided that one has access to a pre-trained diffusion model for the data distribution, one can train a VAE, by training only an encoder by optimizing a novel lower-bound on the data likelihood, which we derived. Our technique outperforms the traditional $\beta$-VAE, producing clear and consistent reconstructions, free of the blurriness typically associated with the latter.


\bibliography{main}
\bibliographystyle{dinat}

\newpage
\appendix

\section{Appendix}

\subsection{Full derivation of the training objective}
\label{sec:mle_objective}
In this section we derive the maximum likelihood training objective for the encoder network.
Let $s_\theta(\textbf{x}_t, t)$ be a score function of a pre-trained unconditional diffusion model and let $e_\phi: (\textbf{x}_t, t) \mapsto (\mu_\phi^z(\textbf{x}_t, t), \sigma^z_\phi(\textbf{x}_t, t)I)$ be the encoder network.
The neural approximation of the conditional data score is given by
\begin{gather*}
s_{\theta, \phi}(\textbf{x}_t, \textbf{z}, t) : = s_\theta(\textbf{x}_t, t) + \nabla_{\textbf{x}_t} \ln q_{t, \phi}(\textbf{z} | \textbf{x}_t)    
\end{gather*}
where $q_{t, \phi}(\textbf{z} | \textbf{x}_t) = \mathcal{N}\big( \textbf{z} ; \mu_\phi^z(\textbf{x}_t, t), \sigma^z_\phi(\textbf{x}_t, t)I \big)$.

Recall that by variational lower bound, for any $\textbf{x}_0, \textbf{z}$ and distribution $q(\textbf{z} | \textbf{x}_0)$ we have
\begin{equation}
     \ln p_{\theta, \phi}(\textbf{x}_0) 
    \geq \mathbb{E}_{\textbf{z} \sim q(\textbf{z} | \textbf{x}_0)}[\ln p_{\theta, \phi}(\textbf{x}_0 | \textbf{z})] - D_{KL}( q(\textbf{z} | \textbf{x}_0) \parallel p(\textbf{z})) 
\end{equation}
Moreover, by \cite[Theorem 3]{song2021maximum} for any $\textbf{x}_0$ and $\textbf{z}$ we have
\begin{align}
 \ln p_{\theta, \phi}(\textbf{x}_0 | \textbf{z})  \geq \mathcal{L}_{DSM}(\textbf{x}_0, \textbf{z})
\end{align}
where
\begin{gather*}
    \mathcal{L}_{DSM}(\textbf{x}_0, \textbf{z}) :=
    \mathbb{E}_{\textbf{x}_t \sim p_t(\textbf{x}_t | \textbf{x}_0)}[\ln \pi(\textbf{x}_t)]  
    -   \mathbb{E}_{\subalign{t \sim U(0,T) \\ \textbf{x}_t \sim p_t(\textbf{x}_t | \textbf{x}_0) }} 
    \big[
    g(t)^2 \norm{ \nabla_{\textbf{x}_t}{\ln{p_t(\textbf{x}_t | \textbf{x}_0)}} - s_{\theta, \phi}(\textbf{x}_t, \textbf{z}, t)}_2^2  
    \big] \\
    + \mathbb{E}_{\subalign{t \sim U(0,T) \\ \textbf{x}_t \sim p_t(\textbf{x}_t | \textbf{x}_0) }} 
    \big[
    g(t)^2 \norm{ \nabla_{\textbf{x}_t}{\ln{p_t(\textbf{x}_t | \textbf{x}_0)}}}_2^2   + 2 \nabla_{\textbf{x}_t} \cdot \textbf{f}(\textbf{x}_t, t) 
    \big] 
\end{gather*}
Putting both together we obtain:
\begin{align*}
    \ln p_{\theta, \phi}(\textbf{x}_0) 
    &\geq \mathbb{E}_{\textbf{z} \sim q_{0,\phi}(\textbf{z} | \textbf{x}_0)}[\ln p_{\theta, \phi}(\textbf{x}_0 | \textbf{z})] -  D_{KL}( q_{0,\phi}(\textbf{z} | \textbf{x}_0) \parallel p(\textbf{z}))  \\
    &\geq \mathbb{E}_{\textbf{z} \sim q_{0,\phi}(\textbf{z} | \textbf{x}_0)}[\mathcal{L}_{DSM}(\textbf{x}_0, \textbf{z})] -  D_{KL}( q_{0,\phi}(\textbf{z} | \textbf{x}_0)  \parallel p(\textbf{z}))  
\end{align*}

after removing terms which don't depend on the parameters of the model and taking average over the data-points, we obtain the following training objective
\begin{equation*}
\begin{aligned}
    \mathcal{L}(\phi)  := \mathbb{E}_{\textbf{x}_0 \sim p(\textbf{x}_0)} \bigg[ 
    \frac{1}{2} \mathbb{E}_{\subalign{t \sim U(0,T) \\ \textbf{x}_t \sim p_t(\textbf{x}_t | \textbf{x}_0) \\ \textbf{z} \sim q_{0,\phi}(\textbf{z} | \textbf{x}_t) }} 
    \big[
    g(t)^2  \norm{ \nabla_{\textbf{x}_t}{\ln{p_t(\textbf{x}_t | \textbf{x}_0)}} - s_{\theta, \phi}(\textbf{x}_t, \textbf{z}, t)}_2^2  
    & \big] \\
    + D_{KL}\big( q_{0,\phi}(\textbf{z} | \textbf{x}_0)  \parallel p(\textbf{z})\big)  & \bigg]
\end{aligned}    
\end{equation*}
It follows from the above considerations that
\begin{gather*}
    \argmin_\phi \mathcal{L} (\phi) = \argmax_\phi \mathbb{E}_{\textbf{x} \sim p(\textbf{x})} [\ln p_{\theta, \phi}(\textbf{x}) ]
\end{gather*}
or in other words that minimizing the objective $\mathcal{L}(\phi) $ is equivalent to maximizing the marginal data log-likelihood $\ln p_{\theta, \phi}(\textbf{x}) $.

Finally just like in $\beta$-VAE we find that in practice it is helpful to introduce a hyper-parameter $\beta \in [0,1]$ which controls the strength of KL regularization. We define our final training objective as:
\begin{equation*}
\begin{aligned}
    \mathcal{L}_\beta(\phi)  := \mathbb{E}_{\textbf{x}_0 \sim p(\textbf{x}_0)} \bigg[ 
    \frac{1}{2} \mathbb{E}_{\subalign{t \sim U(0,T) \\ \textbf{x}_t \sim p_t(\textbf{x}_t | \textbf{x}_0) \\ \textbf{z} \sim q_{0,\phi}(\textbf{z} | \textbf{x}_t) }} 
    \big[
    g(t)^2  \norm{ \nabla_{\textbf{x}_t}{\ln{p_t(\textbf{x}_t | \textbf{x}_0)}} - s_{\theta, \phi}(\textbf{x}_t, \textbf{z}, t)}_2^2  
    & \big] \\
    + \beta D_{KL}\big( q_{0,\phi}(\textbf{z} | \textbf{x}_0)  \parallel p(\textbf{z})\big)  & \bigg]
\end{aligned}    
\end{equation*}

\subsection{Pixel-wise $L_2$ is not a perceptual metric}
\label{sec: l2}
\begin{figure}[h!]
    \centering
    \includegraphics[width=.58\textwidth]{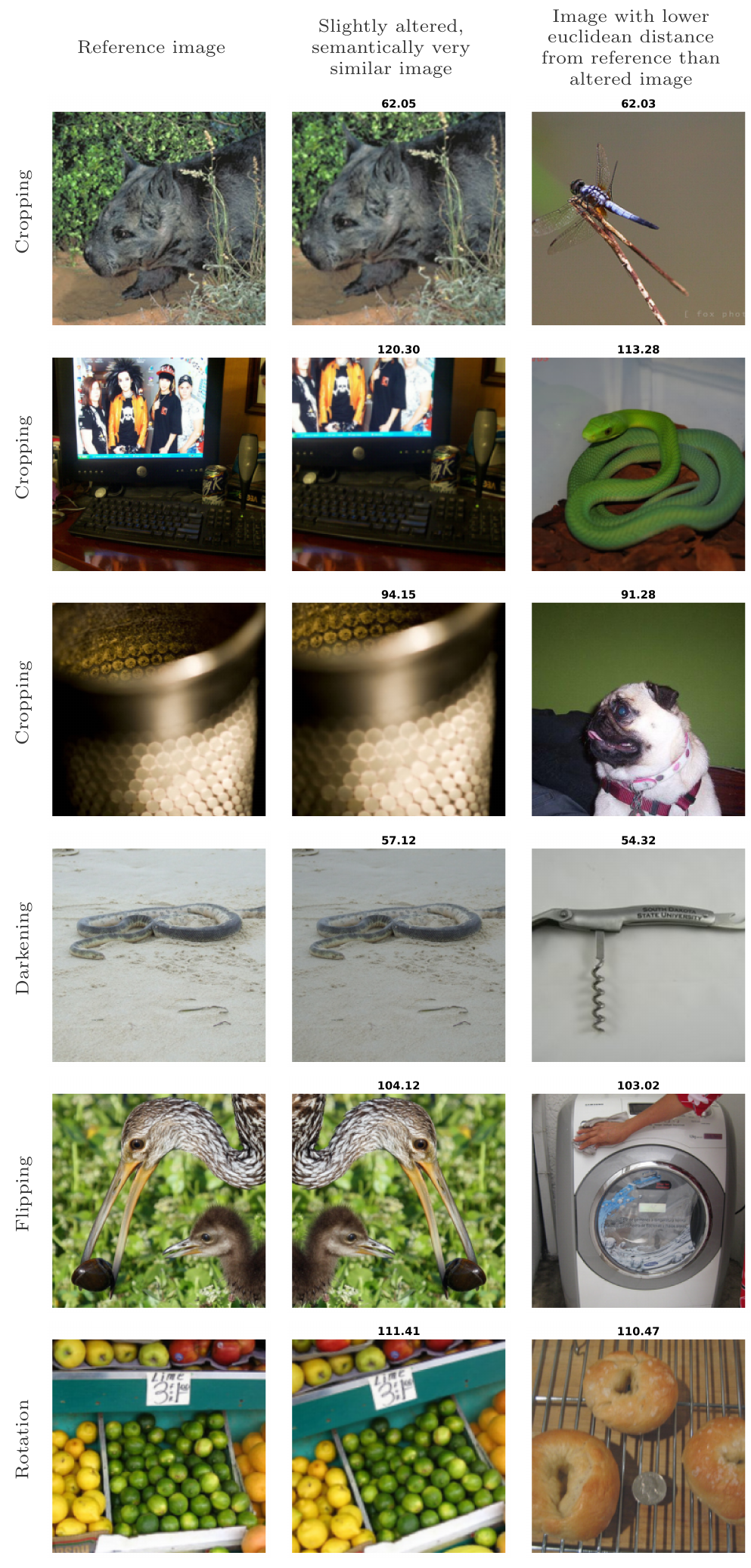}
    \caption{Examples where the $L_2$-distance is not a (semantically) meaningful distance between images. In the left column, a reference image from ImageNet \citep{imagenet} is shown. In the middle column, a slight alteration is applied, whose result a human observer would consider to be very close to the reference image. In the right column, another image from the ImageNet data set is displayed, which to the human observer is very different from the reference image, but which has a lower $L_2$-distance to the reference image than the altered image. The numbers above the images indicate the $L_2$-distance to the reference image. Figure taken from \cite{stanczuk2021wasserstein}.}
    \label{fig:L2bad}
\end{figure}

As we discussed in the previous section choosing a Gaussian model for $ p(\textbf{x} | \textbf{z}) $ in VAEs leads to a term in the loss function, which is equivalent to the $L_2$ reconstruction error. While $L_2$ is a common dissimilarity metric it is a very bad choice for certain data modalities such as images. This is because $L_2$ distance measures the differences between corresponding pixel intensities, but does not take into account human perception. Thus, two images may have a small $L_2$ distance but still appear visually different, or vice versa (see Figure \ref{fig:L2bad}). The metric does not consider the hierarchical and contextual information that humans use when perceiving images. In particular small spatial shifts, rotations or cropping can lead to large $L_2$ distances even if the images are perceptually similar.

\section{Experimental details} \label{Experimental details}

\subsection{ScoreVAE}
The pretrained diffusion models for all experiments are based on the DDPM architecture \cite{ddpm}. We used $128$ base filters and attention at resolution $16\times 16$ for all experiments. For Cifar10, we set the channel multiplier array to $[1, 2, 2, 2]$ and the number of ResNet blocks to $4$. For CelebA $64\times 64$, we set the channel multiplier array to $[1, 1, 2, 2, 3]$ and the number of ResNet blocks to $2$. We used $0.1$ dropout rate for Cifar10 and $0$ dropout rate for CelebA. We used the beta-linear variance preserving forward process with the same parameters as the ones used by \cite{song2020score} and trained the diffusion model using the weighted denoising score matching objective with the simple weighting, i.e., $\lambda(t)=\sigma(t)^2$, where $\sigma(t)$ is the standard deviation of the perturbation kernel. We used the Adam optimizer and EMA rate 0.999. Finally, we set the learning rate to $2\mathrm{e}{-4}$ for Cifar10 and $1\mathrm{e}{-4}$ for CelebA.

The time dependent encoder for the Cifar10 experiment is a simple convolutional network that consists of a sequence of blocks of convolutions followed by the GELU activation function. The final activation is flattened and concatenated to the time tensor. A final linear layer maps the time augmented flattened tensor to the latent dimension. The time dependent encoder for CelebA is based heavily on the DDPM architecture. We removed the upsampling part of the U-NET and removed the skip connections. The downscaled tensor is flattened and mapped to the latent dimension with an additional linear layer. We used the Adam optimizer and EMA rate 0.999. We set the learning rate to $2\mathrm{e}{-4}$ for Cifar10 and $1\mathrm{e}{-4}$ for CelebA. We trained the Cifar10 encoder for $1.4M$ iterations and the CelebA encoder for $300K$ iterations.

\subsection{VAE}

For VAE we used exactly the same encoder architectures as in the Score VAE (except they were not conditioned on time). For each choice of encoder we created a mirror decoder with symmetric architecture. In Cifar10 the decoder starts by reshaping the the flat latent vector into a tensor which is then passed through a sequence of transposed convolutions which exactly mirror the structure of the encoder. In CelebA we used a decoder consisting of the upsampling part of the DDPM U-NET. The Cifar10 model was trained for 11M iterations, while the CelebA model was trained for 600K iterations.

\newpage
\section{Extended qualitative evaluation}\label{sec:extended qualitative evaluation}

\begin{table}[h!]
    \centering
    \resizebox{0.9\textwidth}{!}{
    \begin{tabular}{cccccc}
        Original & \multicolumn{1}{c}{ScoreVAE} & \multicolumn{1}{c}{ScoreVAE+} & \multicolumn{2}{c}{DiffDecoder} & \multicolumn{1}{c}{VAE}  \\ 
        & ($\beta=0.01$) & ($\beta=0.01$) & ($\beta=0.01$) & ($\beta=0$) & ($\beta=0.01$) \\

        \includegraphics[width=.145\textwidth]{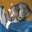} &   
        \includegraphics[width=.145\textwidth]{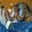} &
        \includegraphics[width=.145\textwidth]{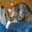} &
        \includegraphics[width=.145\textwidth]{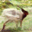} &
        \includegraphics[width=.145\textwidth]{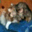} &
        \includegraphics[width=.145\textwidth]{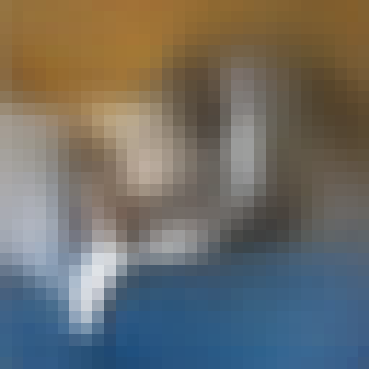} \\

        \includegraphics[width=.145\textwidth]{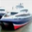} &   
        \includegraphics[width=.145\textwidth]{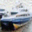} &
        \includegraphics[width=.145\textwidth]{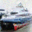} &
        \includegraphics[width=.145\textwidth]{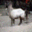} &
        \includegraphics[width=.145\textwidth]{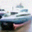} &
        \includegraphics[width=.145\textwidth]{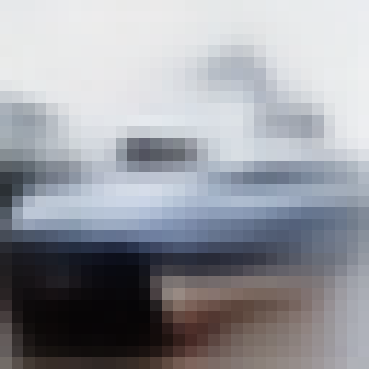} \\
 
        \includegraphics[width=.145\textwidth]{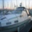} &   
        \includegraphics[width=.145\textwidth]{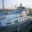} &
        \includegraphics[width=.145\textwidth]{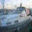} &
        \includegraphics[width=.145\textwidth]{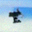} &
        \includegraphics[width=.145\textwidth]{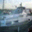} &
        \includegraphics[width=.145\textwidth]{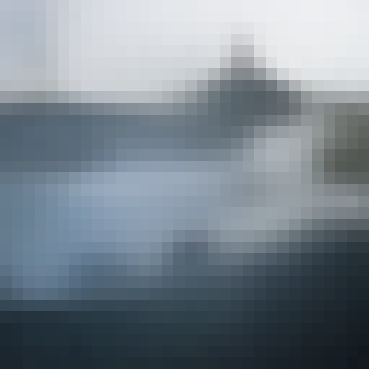} \\

        \includegraphics[width=.145\textwidth]{figures/images/cifar10/original/4.png} &   
        \includegraphics[width=.145\textwidth]{figures/images/cifar10/reconstruction/4.png} &
        \includegraphics[width=.145\textwidth]{figures/images/cifar10/corrected_reconstruction/4.png} &
        \includegraphics[width=.145\textwidth]{figures/images/cifar10/diffusion_decoder_beta_0.01/4.png} &
        \includegraphics[width=.145\textwidth]{figures/images/cifar10/diffusion_decoder_beta_0/4.png} &
        \includegraphics[width=.145\textwidth]{figures/images/cifar10/VAE_reconstruction/4.png} \\
        \includegraphics[width=.145\textwidth]{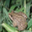} &   
        \includegraphics[width=.145\textwidth]{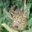} &
        \includegraphics[width=.145\textwidth]{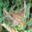} &
        \includegraphics[width=.145\textwidth]{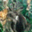} &
        \includegraphics[width=.145\textwidth]{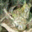} &
        \includegraphics[width=.145\textwidth]{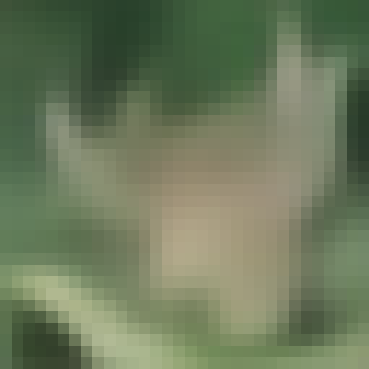} \\
        \includegraphics[width=.145\textwidth]{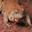} &   
        \includegraphics[width=.145\textwidth]{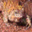} &
        \includegraphics[width=.145\textwidth]{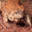} &
        \includegraphics[width=.145\textwidth]{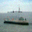} &
        \includegraphics[width=.145\textwidth]{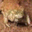} &
        \includegraphics[width=.145\textwidth]{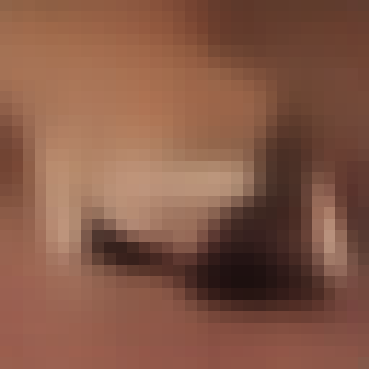} \\
        \includegraphics[width=.145\textwidth]{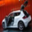} &   
        \includegraphics[width=.145\textwidth]{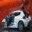} &
        \includegraphics[width=.145\textwidth]{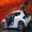} &
        \includegraphics[width=.145\textwidth]{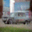} &
        \includegraphics[width=.145\textwidth]{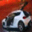} &
        \includegraphics[width=.145\textwidth]{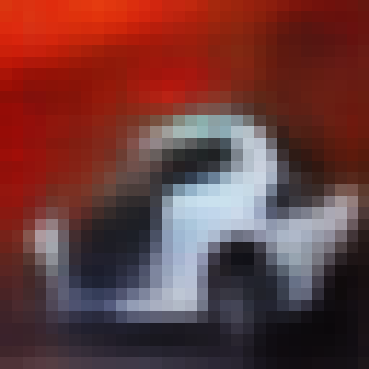} \\
        \includegraphics[width=.145\textwidth]{figures/images/cifar10/original/8.png} &   
        \includegraphics[width=.145\textwidth]{figures/images/cifar10/reconstruction/8.png} &
        \includegraphics[width=.145\textwidth]{figures/images/cifar10/corrected_reconstruction/8.png} &
        \includegraphics[width=.145\textwidth]{figures/images/cifar10/diffusion_decoder_beta_0.01/8.png} &
        \includegraphics[width=.145\textwidth]{figures/images/cifar10/diffusion_decoder_beta_0/8.png} &
        \includegraphics[width=.145\textwidth]{figures/images/cifar10/VAE_reconstruction/8.png} \\
        \includegraphics[width=.145\textwidth]{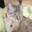} &   
        \includegraphics[width=.145\textwidth]{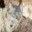} &
        \includegraphics[width=.145\textwidth]{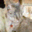} &
        \includegraphics[width=.145\textwidth]{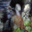} &
        \includegraphics[width=.145\textwidth]{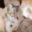} &
        \includegraphics[width=.145\textwidth]{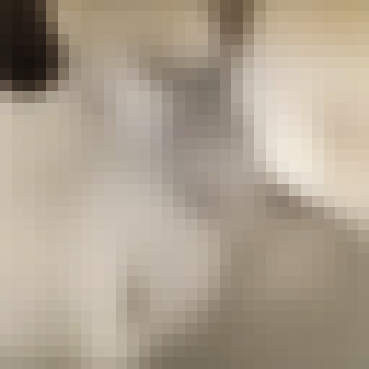} \\
        \includegraphics[width=.145\textwidth]{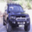} &   
        \includegraphics[width=.145\textwidth]{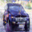} &
        \includegraphics[width=.145\textwidth]{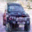} &
        \includegraphics[width=.145\textwidth]{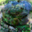} &
        \includegraphics[width=.145\textwidth]{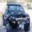} &
        \includegraphics[width=.145\textwidth]{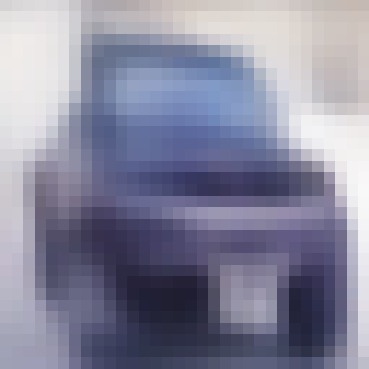} \\
        \includegraphics[width=.145\textwidth]{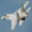} &   
        \includegraphics[width=.145\textwidth]{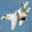} &
        \includegraphics[width=.145\textwidth]{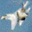} &
        \includegraphics[width=.145\textwidth]{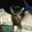} &
        \includegraphics[width=.145\textwidth]{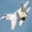} &
        \includegraphics[width=.145\textwidth]{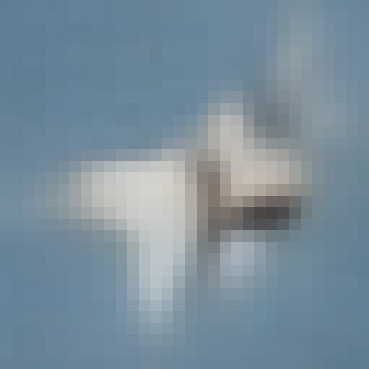} \\
    \end{tabular}}
    \caption{Cifar10}
    \label{fig:extended cifar10 qualitative comparison}
\end{table}

\begin{table}[h!]
    \centering
    \resizebox{0.9\textwidth}{!}{
    \begin{tabular}{cccccc}
        Original & \multicolumn{1}{c}{ScoreVAE} & \multicolumn{1}{c}{ScoreVAE+} & \multicolumn{2}{c}{DiffDecoder} & \multicolumn{1}{c}{VAE}  \\ 
        & ($\beta=0.01$) & ($\beta=0.01$) & ($\beta=0.01$) & ($\beta=0$) & ($\beta=0.01$) \\

        \includegraphics[width=.145\textwidth]{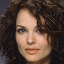} &   
        \includegraphics[width=.145\textwidth]{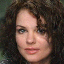} &
        \includegraphics[width=.145\textwidth]{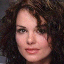} &
        \includegraphics[width=.145\textwidth]{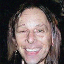} &
        \includegraphics[width=.145\textwidth]{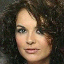} &
        \includegraphics[width=.145\textwidth]{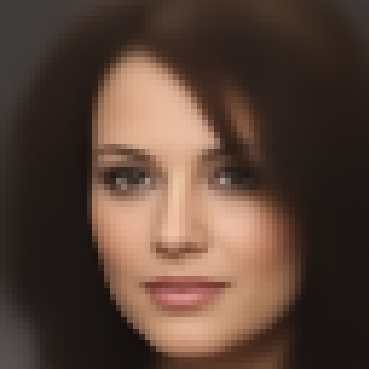} \\

        \includegraphics[width=.145\textwidth]{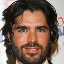} &   
        \includegraphics[width=.145\textwidth]{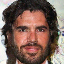} &
        \includegraphics[width=.145\textwidth]{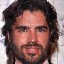} &
        \includegraphics[width=.145\textwidth]{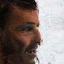} &
        \includegraphics[width=.145\textwidth]{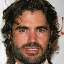} &
        \includegraphics[width=.145\textwidth]{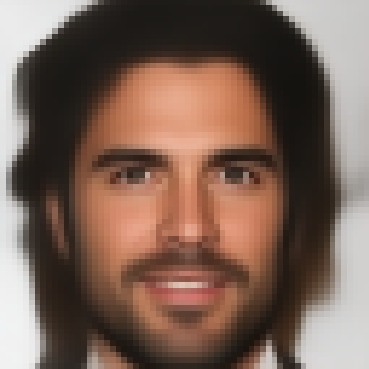} \\
 
        \includegraphics[width=.145\textwidth]{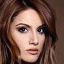} &   
        \includegraphics[width=.145\textwidth]{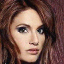} &
        \includegraphics[width=.145\textwidth]{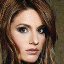} &
        \includegraphics[width=.145\textwidth]{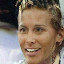} &
        \includegraphics[width=.145\textwidth]{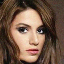} &
        \includegraphics[width=.145\textwidth]{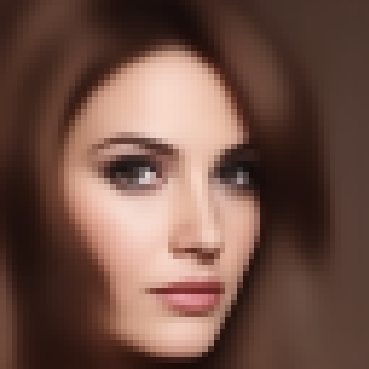} \\

        \includegraphics[width=.145\textwidth]{figures/images/celebA/original/4.png} &   
        \includegraphics[width=.145\textwidth]{figures/images/celebA/reconstruction/4.png} &
        \includegraphics[width=.145\textwidth]{figures/images/celebA/corrected_reconstruction/4.png} &
        \includegraphics[width=.145\textwidth]{figures/images/celebA/diffusion_decoder_beta_0.01/4.png} &
        \includegraphics[width=.145\textwidth]{figures/images/celebA/diffusion_decoder_beta_0/4.png} &
        \includegraphics[width=.145\textwidth]{figures/images/celebA/VAE_reconstruction/4.png} \\
        \includegraphics[width=.145\textwidth]{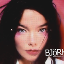} &   
        \includegraphics[width=.145\textwidth]{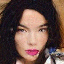} &
        \includegraphics[width=.145\textwidth]{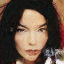} &
        \includegraphics[width=.145\textwidth]{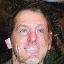} &
        \includegraphics[width=.145\textwidth]{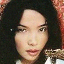} &
        \includegraphics[width=.145\textwidth]{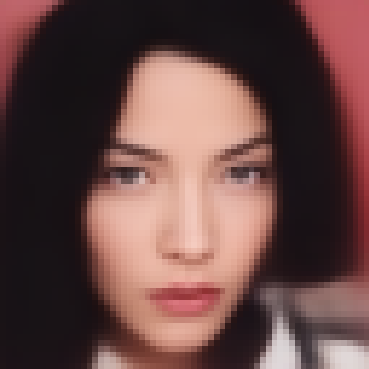} \\
        \includegraphics[width=.145\textwidth]{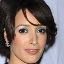} &   
        \includegraphics[width=.145\textwidth]{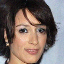} &
        \includegraphics[width=.145\textwidth]{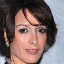} &
        \includegraphics[width=.145\textwidth]{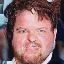} &
        \includegraphics[width=.145\textwidth]{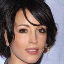} &
        \includegraphics[width=.145\textwidth]{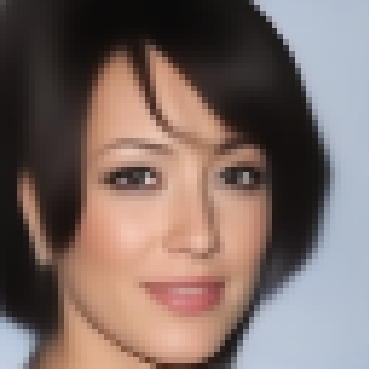} \\
        \includegraphics[width=.145\textwidth]{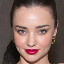} &   
        \includegraphics[width=.145\textwidth]{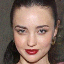} &
        \includegraphics[width=.145\textwidth]{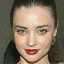} &
        \includegraphics[width=.145\textwidth]{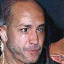} &
        \includegraphics[width=.145\textwidth]{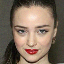} &
        \includegraphics[width=.145\textwidth]{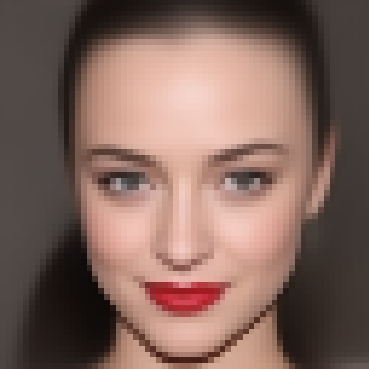} \\
        \includegraphics[width=.145\textwidth]{figures/images/celebA/original/8.png} &   
        \includegraphics[width=.145\textwidth]{figures/images/celebA/reconstruction/8.png} &
        \includegraphics[width=.145\textwidth]{figures/images/celebA/corrected_reconstruction/8.png} &
        \includegraphics[width=.145\textwidth]{figures/images/celebA/diffusion_decoder_beta_0.01/8.png} &
        \includegraphics[width=.145\textwidth]{figures/images/celebA/diffusion_decoder_beta_0/8.png} &
        \includegraphics[width=.145\textwidth]{figures/images/celebA/VAE_reconstruction/8.png} \\
        \includegraphics[width=.145\textwidth]{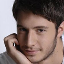} &   
        \includegraphics[width=.145\textwidth]{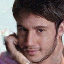} &
        \includegraphics[width=.145\textwidth]{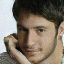} &
        \includegraphics[width=.145\textwidth]{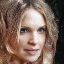} &
        \includegraphics[width=.145\textwidth]{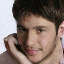} &
        \includegraphics[width=.145\textwidth]{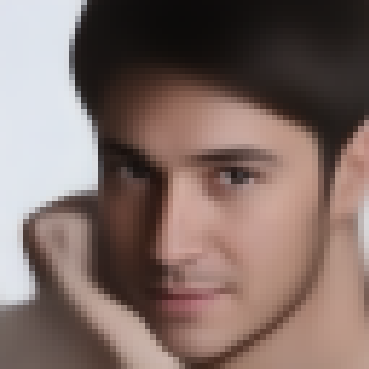} \\
        \includegraphics[width=.145\textwidth]{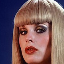} &   
        \includegraphics[width=.145\textwidth]{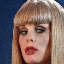} &
        \includegraphics[width=.145\textwidth]{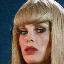} &
        \includegraphics[width=.145\textwidth]{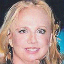} &
        \includegraphics[width=.145\textwidth]{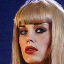} &
        \includegraphics[width=.145\textwidth]{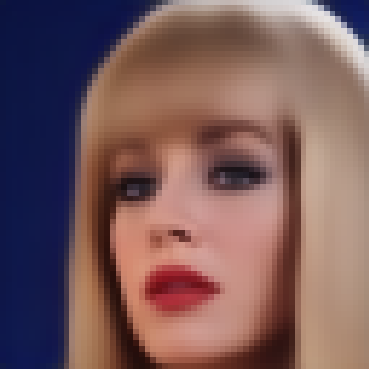} \\
        \includegraphics[width=.145\textwidth]{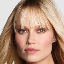} &   
        \includegraphics[width=.145\textwidth]{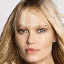} &
        \includegraphics[width=.145\textwidth]{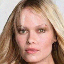} &
        \includegraphics[width=.145\textwidth]{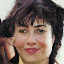} &
        \includegraphics[width=.145\textwidth]{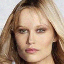} &
        \includegraphics[width=.145\textwidth]{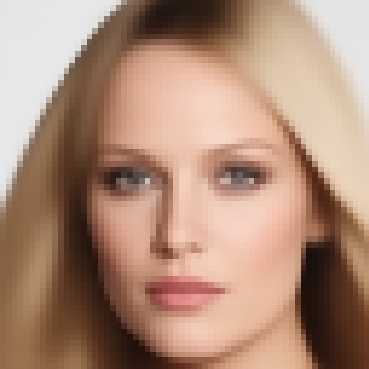} \\
    \end{tabular}}
    \caption{CelebA}
    \label{fig:extended celebA qualitative comparison}
\end{table}

\end{document}